# TBBC: Predict True Bacteraemia in Blood Cultures via Deep Learning

Kira Sam*

*Datalink Research and Technology Lab

**Abstract**

Bacteraemia is a blood stream infection with a high morbidity and mortality rate. Accurately diagnosing bacteraemia using blood cultures is a resource-intensive process. Developing a machine learning model to predict the outcome of a blood culture in the emergency department has the potential to improve diagnosis and reduce healthcare costs and mitigate antibiotic use. This thesis aims to identify machine learning techniques to predict bacteraemia and develop a predictive model using data from the emergency department of St. Antonius Hospital. Based on current literature, CatBoost and random forest were selected as the most promising machine learning techniques for bacteraemia prediction. Model optimisation using Optuna focused on maximising sensitivity to accurately identify patients with bacteraemia. The final random forest model achieved an ROC AUC of 0.78 and demonstrated a sensitivity of 0.92 on the test set. Notably, the model could accurately identify patients that had a low risk of bacteraemia at 36.02%, at the cost of 0.85% false negatives. Based on these findings, implementing the model into the emergency department at St. Antonius Hospital could reduce the number of blood cultures taken as well as lowering healthcare costs and antibiotic treatments. Further studies could focus on externally validating the model, exploring advanced machine learning techniques and removing potential confounders in the data set to ensure the model's generalisability.

## 1. Introduction

The World Health Organization (WHO) considers rising antimicrobial resistance a global concern [1]. Antimicrobial resistance reduces treatment efficacy and increases disease spread and mortality rates [1]–[3]. Bacteraemia, bacteria in the bloodstream, is a severe condition with high morbidity and mortality [4], [5]. Blood cultures (BC) are the gold-standard diagnostic test, but results take hours to days, leading to early antibiotic prescription [6]. This yields low true positives and high contamination rates [7]–[10]. Each culture costs up to 250 euros [11], and misdiagnoses lead to prolonged admissions and additional medication [12].

Machine learning can improve diagnostic accuracy and efficiency [14]. Predictive models can identify low-risk patients, reducing unnecessary antibiotic use and BC testing [10], [15]. Studies have applied machine learning to predict true bacteraemia (TB) [15], [19]–[26], including the Shapiro [27], MPB-INFURG-SEMES [23], and VUMC models [20]. However, implementation and validation challenges persist due to hospital-specific features and patient population variability [19], [23].

## 2. Literature Review

Predicting true bacteraemia (TB) using machine learning has shown promise [15], [19]–[26]. The Shapiro model achieved a ROC AUC of 0.80 and reduced blood culture (BC) usage by 27% [27]. However, its performance varied across studies [23]. Boerman et al.'s (2022) logistic regression and XGBoost models achieved ROC AUCs of 0.78 and 0.77, respectively [20]. Schinkel et al.'s (2022) implementation study yielded ROC AUCs of 0.81, 0.80, 0.76, and 0.75 across multiple hospitals [19].

Other studies have also demonstrated effective TB prediction. Roimi et al. (2020) achieved ROC AUCs of 0.87-0.93 using ensemble models [21]. Garnica et al. (2021) developed classifiers with ROC AUCs up to 0.93 [15]. Julián-Jiménez et al. (2021) built a risk model with a ROC AUC of 0.924 [23]. Lee et al. (2022) constructed a multilayer feedforward neural network with a ROC AUC of 0.762 [25]. Choi et al. (2022) developed XGB models with ROC AUCs of 0.718 and 0.853 [24]. Chang et al. (2023) utilized cell population data and achieved ROC AUCs up to 0.844 [22]. McFadden et al. (2023) trained models with ROC AUCs of 0.76-0.82 [26].

XGBoost was frequently used, but Categorical Boosting (CB) and Random Forest (RF) show potential due to their unique advantages [22], [31], [33]. This study will explore CB and RF for TB prediction. This literature review highlights various machine learning approaches for TB prediction. CB and RF demonstrate promise due to their advantages. This study aims to provide new insights into TB prediction using CB and RF.

## 3. Methodology

This study collected electronic health records (EHR) from St. Antonius Hospital's emergency department (ED) between January 2018 and July 2023. The dataset consisted of 27,009 adult patients with blood cultures (BC) taken. Patients with neutropenia were excluded. The dataset included 25 categorical and 25 continuous variables, along with BC results (negative/positive) [34].

### 3.1. Evaluation Metrics

To evaluate true bacteraemia (TB) prediction models, this study used:

- Sensitivity (True Positive Rate)
- Specificity
- Accuracy
- Precision (Positive Predictive Value)
- F1 Score
- Area Under the Receiver Operating Characteristic Curve (ROC AUC)
- Area Under the Precision-Recall Curve (PR AUC)

### 3.2. CatBoost, Random Forest, and Optuna

This study employed CatBoost (CB) for handling categorical features and missing data [35] Random Forest (RF) for robust prediction accuracy [33] Optuna for hyperparameter optimization using Bayesian optimization [36]

### 3.3. Model Development

The dataset was split into training (80%), testing (10%), and validation (10%) sets. Optuna tuned hyperparameters for CB and RF models, focusing on maximizing sensitivity. Class weights were added to reward minority class detection.

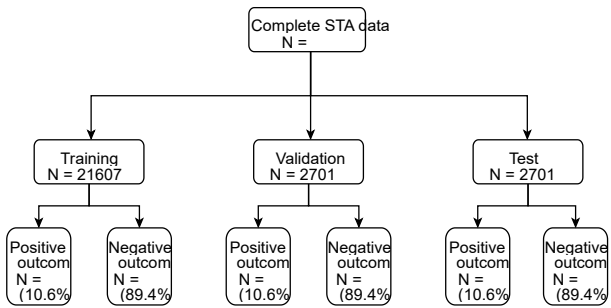

**Fig 1**. Overview of data sets after splitting.

### 3.4. Statistical Analysis

A Mann-Whitney U test compared continuous variable distributions between positive and negative BC outcome groups (p-value < 0.05). Results are presented in Appendix 7.5.

### 3.5. Ethical Considerations

Data was pseudonymised, and patients could opt out. The Medical Ethics Review Committee waived formal approval and informed consent (reference Z23.042). Data was stored securely, and analysis was conducted within the hospital's protected workspace.

## 4. Result

The dataset consisted of 27,009 blood culture (BC) samples from St. Antonius Hospital's emergency department (ED) between January 2018 and July 2023. Positive BC samples accounted for 10.6% (n=2,872), while negative samples accounted for 89.4% (n=24,138). The median age was 69 (IQR 55-78), with 45.3% female patients.

### 4.1. Model Performances

The performance metrics of the final CatBoost (CB) and Random Forest (RF) models. **Table 01**

RF outperformed CB in accuracy, sensitivity, specificity, precision, F1 score, ROC AUC, and PR AUC.

**Table 2.** Performance metrics of the final CatBoost and Random Forest models.

| Metric | CatBoost | Random Forest |
|---|---|---|
| Accuracy | 0.416 | 0.458 |
| Sensitivity | 0.916 | 0.920 |
| Specificity | 0.357 | 0.403 |
| Precision | 0.145 | 0.155 |
| F1 Score | 0.250 | 0.265 |
| ROC AUC | 0.767 | 0.782 |
| PR AUC | 0.304 | 0.342 |

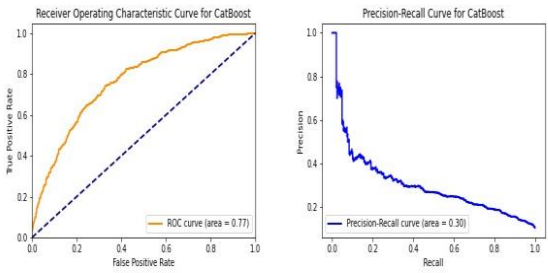

**Fig 2.** ROC AUC and PR AUC of final CatBoost model.

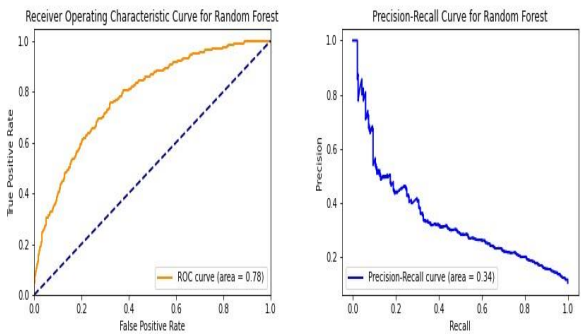

**(a)** ROC AUC        **(b)** PR AUC

**Fig 3.** ROC AUC and PR AUC of final random forest model.

### 4.3. Confusion Matrices and Thresholds

The confusion matrices and histograms for both models are presented. The optimal thresholds for CB and RF were 0.4 and 0.3, respectively.

#### 4.3.1. CatBoost Model

The CB model predicts that with the threshold set at 0.4, 861 (31.9% of thetotal) of the BC can be withheld, at the expense of 24 false negatives (0.89%). The confusion matrix in visually represents these predictions, illustrating the threshold effect at 0.4

#### 4.3.2. Random Forest Model

The RF model, at a threshold of 0.3, identifies 973 cases (36.02%) where the BC can be omitted, with 23 false negatives (0.85%), analysis by illustrating the probability distribution and the impact of setting a threshold at 0.3.

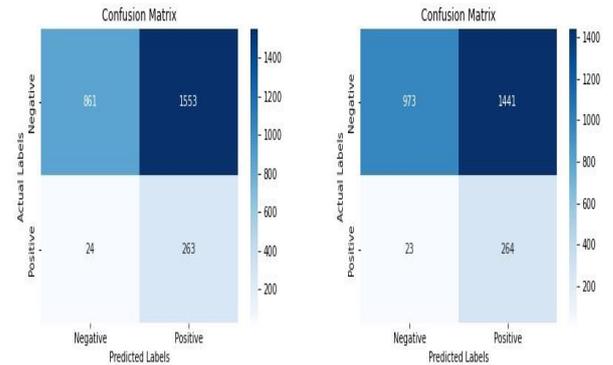

**(a)** CatBoost - threshold = 0.4 **(b)** Random forest - threshold = 0.3

**Fig 4.** Comparison of confusion matrices for CatBoost and random forest models.

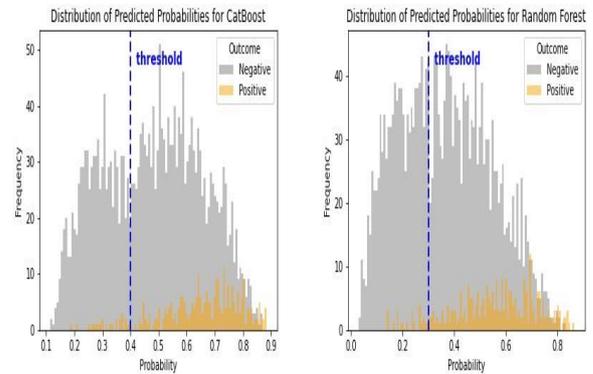

**(a)** CatBoost - threshold = 0.4 **(b)** Random forest - threshold = 0.3

**Fig 5.** Comparison of probability thresholds for CatBoost and random forest models.

### 4.4 Feature Importance

SHAP plots highlighted the top 20 predictors for predicting true bacteraemia (TB). Lymphocytes were the most influential predictor for both CB and RF. Other notable predictors included bilirubin, neutrophils, urea, eosinophils, and temperature. The ROC AUC, PR AUC, confusion matrices, probability thresholds, and SHAP plots for both models.

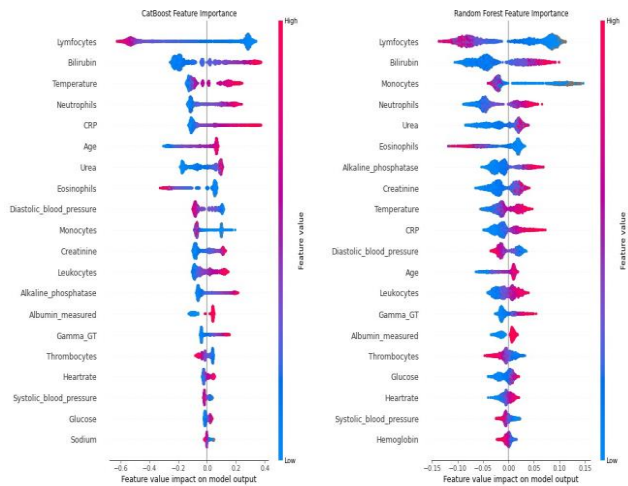

**(a)** CatBoost    **(b)** Random forest

**Fig 6:** SHAP-plots of the feature importance of the CatBoost and random forest models, showing the 20 most important variables for predicting TB.

## 5. Discussion

This paper aimed to identify the most promising machine learning technique for predicting true bacteraemia (TB) and develop a sensitive TB prediction model using CatBoost (CB) and Random Forest (RF) on St. Antonius Hospital's emergency department electronic health records.

### 5.1. Validity of the Study and Methods

The study's data set, consisting of 27,009 blood culture samples, may have introduced bias towards bacteraemia due to clinical suspicion influencing selection criteria. However, the use of Optuna for hyperparameter tuning and stratified data split ensured reliability and robustness.

### 5.2. Interpretation of Results

Both CB and RF demonstrated promising predictive capabilities, with ROC AUCs of 0.77 and 0.78, respectively. RF outperformed CB in sensitivity, specificity, and true negative rate.

### 5.2.1. Model Performance Evaluation

RF's higher true negative rate suggests its effectiveness in correctly identifying patients without bacteraemia, making it the best model for TB prediction. Its robust performance, resistance to overfitting, and internal handling of missing data contribute to its suitability.

### 5.2.2. Random Forest Interpretation

Comparing RF models, this study's RF achieved higher sensitivity and lower false negatives (0.85%) than Garnica et al. (2021) and McFadden et al. (2023) [15, 26].

**Table 3:** Performance metrics comparison of random forest models. (TN = True Negatives, FN = False Negative)

| Metric | This study's RF | Garnica et al. | McFadden et al. |
|---|---|---|---|
| Accuracy | 0.458 | 0.859 | 0.694 |
| Sensitivity | 0.92 | 0.874 | 0.664 |
| Specificity | 0.403 | 0.844 | 0.723 |
| ROC AUC | 0.78 | 0.93 | 0.82 |
| TN % of Total | 36.02% | 42.1% | 39.3% |
| FN % of Total | 0.85% | 6.31% | 15.33% |
| Total Test Set | 2701 | 871 | 3875 |

### 5.2.3. CatBoost Interpretation

Although CB did not outperform RF, it achieved a lower number of false negatives (0.89%) compared to Chang et al. (2022) and Bopche et al. (2024) [22, 32].

### 5.2.4. Comparison with VUMC Model

This study's RF model performed better than the VUMC model and STA validation, with a lower percentage of unnecessary blood cultures (36.02%) and false negatives (0.85%).

**Table 4:** Performance metrics comparison of CatBoost models. (TN = True Negatives, FN = False Negative)

| Metric | This study's CB | Chang et al. | Bopche et al. |
|---|---|---|---|
| Accuracy | 0.416 | 0.844 | 0.853 |
| Sensitivity | 0.916 | 0.715 | 0.627 |
| Specificity | 0.357 | 0.826 | 0.875 |
| ROC AUC | 0.77 | 0.84 | 0.82 |
| TN % of Total | 31.9% | 74.7% | 79.7% |
| FN % of Total | 0.89% | 2.71% | 3.32% |
| Total Test Set | 2701 | 3143 | 13195 |

Table 5: Performance metrics comparison of this study, the VUMC model and the STA validation. (TN = True Negatives, FN = False Negative)

| Metric | This study's CB | This study's RF | Boerman et al. | STA validation |
|---|---|---|---|---|
| Accuracy | 0.416 | 0.458 | 0.481 | 0.447 |
| Sensitivity | 0.916 | 0.920 | 0.916 | 0.925 |
| Specificity | 0.416 | 0.403 | 0.422 | 0.39 |
| ROC AUC | 0.77 | 0.78 | 0.77 | 0.79 |
| TN % of Total | 31.9% | 36.02% | 37.1% | 34.9% |
| FN % of Total | 0.89% | 0.85% | 1.02% | 0.79% |
| Total Test Set | 2701 | 2701 | 1277 | 27009 |

### 5.2.5. Optional thresholds

This study evaluated the 5% threshold, Therefore, the thresholds were set at 40% for CB and 30% for RF, optimising for the best balance between true and false negatives for each individual model, which is crucial for clinical utility. For both models, optional thresholds of 0.3 for CB and 0.4 for RF. With the threshold at 0.3, the CB model predicts that 489 (18.1%) of the BC can be omitted, at the cost of 9 (0.33%) false negatives. This is a significantly lower number of false negatives, compared to the 0.4 threshold with 0.89%. However, it also reduced the number of true negatives by almost half in comparison to the 0.4 threshold (31.9%). In conclusion, the CB model could further eliminate false negatives, but this would significantly increase the false positives to such a degree that implementation of the model would not be useful in a real clinical setting. Having a threshold of 0.4 for the RF model resulted in higher percentage of true negatives, 51.98% versus the 36.03% with the threshold at 0.3. However, this drastically increased the false negatives rate from 0.85% with the threshold at 0.3, to 1.89%. Adjusting the threshold for the RF would allow for more negatives to be predicted correctly, but at the cost of more than double the false negatives.

Table 6: Performance metrics of the final CatBoost and random forest models with the optional thresholds of 0.3 (CB) and 0.4 (RF). (TN = True Negatives, FN = False Negative)

| Metric | CatBoost | Random forest |
|---|---|---|
| Accuracy | 0.284 | 0.607 |
| Sensitivity | 0.969 | 0.822 |
| Specificity | 0.203 | 0.582 |
| FN Percentage | 0.33% | 1.89% |
| TN Percentage | 18.1% | 51.98% |

### 5.3. Statistical Analysis and Feature Importance

The Mann-Whitney U test revealed significant differences in most variables between outcome groups. SHAP plots highlighted lymphocytes, bilirubin, neutrophils, urea, eosinophils, and temperature as key predictors.

### 6. Limitations

Potential false positives in blood cultures due to contamination [7-10] may lead to unnecessary antibiotic treatment. Patients with central venous lines, prosthetic material, or specific diagnoses were not filtered out, potentially skewing model training. Positive outcomes were the minority class, addressed through class weights in hyperparameters [47).

### 7. Recommendations

Implementation and external validation conduct prospective real-time evaluation and pilot studies to integrate the RF model into hospital systems. Explore ways to mitigate false positives without compromising sensitivity. Identify and remove edge cases, and review false negatives to adjust the RF model's threshold. Upgrade server infrastructure to explore advanced techniques like neural networks. Incorporate feature selection, focusing on significant variables and collaborating with physicians to optimize model effectiveness.

### 8. Conclusion

This paper aimed to find an answer to the question: "How can the most promising machine learning techniques identified from current literature be applied to develop a predictive model for true bacteraemia, and how does this model perform in terms of sensitivity within the emergency department of St. Antonius Hospital?". This was executed by employing a data set of 27009 patients with blood culture results. CatBoost and random forest were chosen as most promising machine learning techniques based on current literature. The final selected model, random forest, achieved a ROC AUC of 0.78, indicating its ability to effectively predict bacteraemia. The model predicted that for 973 (36.02%) of the patients blood cultures could have been withheld, with only 23 missed cultures (0.85%). Implementing this model could lead to significant reductions in unnecessary testing, antibiotic treatments, and related healthcare costs, while maintaining high sensitivity in identifying patients at risk of bacteraemia. It is important to note that while the model may increase false positives, this is mitigated by current clinical practices where blood cultures are routinely ordered for suspected cases. By identifying true cases, this study contributes to the crucial effort of preventing antimicrobial resistance. Future research should focus on validating this model using external data and clinical pilot studies to prepare for real-world implementation. Enhancing the data quality and exploring advanced techniques using updated software could further improve the accuracy and utility of the model. This study aims to improve predictive abilities to help physicians make better decisions and improve patient outcomes in managing bacteraemia.

# Appendix

## A. Continuous Variables Explanations

**Table 7:** Explanation of the continuous variables in the data set

| Variable | Explanation |
| --- | --- |
| Alkaline_phosphatase | Liver enzyme indicator |
| Basophils | Type of white blood cell percentage |
| Bilirubin | Pigment in bile produced by the liver |
| Creatinine | Waste product from muscle metabolism |
| CRP | Marker of inflammation |
| Eosinophils | Type of white blood cell percentage |
| Gamma_GT | Liver enzyme indicator |
| Glucose | Blood sugar level |
| Hemoglobin | Oxygen-carrying protein in red blood cells |
| Hematocrit | Volume percentage of red blood cells in blood |
| Leukocytes | White blood cell count |
| Lymfocytes | Type of white blood cell count |
| Monocytes | Type of white blood cell count |
| Neutrophils | Type of white blood cell count |
| Potassium | Electrolyte level in blood |
| Sodium | Electrolyte level in blood |
| Thrombocytes | Platelet count |
| Urea | Waste product from protein metabolism |
| Heart_rate | Beats per minute of the heart |
| Systolic_blood_pressure | Pressure in arteries when the heart beats |
| Diastolic_blood_pressure | Pressure in arteries when the heart rests |
| Temperature | Body temperature |
| Respiratory_rate | Breaths per minute |
| Saturation | Oxygen saturation level in the blood |

## B. Blood Culture Protocol

A blood culture involves an aerobic and an anaerobic medium containing BHI broth, resin beads, and growth factors. It is incubated for five days, with the incubator periodically measuring the color change of a $CO_2$ indicator at the bottom of the medium. A positive blood culture turns from green to yellow due to $CO_2$ production. Four bottles are collected from the patients: two with a green cap (aerobic bacteria) and two with an orange cap (anaerobic bacteria). These bottles are incubated, with the incubator monitoring $CO_2$ levels. When sufficient growth occurs, the culture becomes 'positive', and the microbiologist notifies the physician to begin empirical treatment. The whole culture is considered to be positive when one or more bottles are positive, as long as the identified bacteria is not part of the list of contaminated bacteria. Identifying the bacteria requires spreading a sample from the bottles on a culture plate for further incubation. Analysts examine these plates daily, and based on the colonies' appearance, they can often identify the bacteria. For confirmation, the spectrometer is used to precisely identify the species and check for antibiotic resistance, facilitating the optimal treatment of the patients.

**C. Hyperparameter Tuning Details**

*a. Hyperparameters used for tuning*

Table 8: Hyperparameters used for tuning with Optuna for CatBoost

| Hyperparameter | Range |
|---|---|
| Learning rate | $1 \times 10^{-3}$ to 0.1 (log scale) |
| Depth | 1 to 10 |
| Subsample | 0.05 to 1.0 |
| Colsample by level | 0.05 to 1.0 |
| Minimum data in leaf | 1 to 100 |
| Class weight (1) | 1 to 12 |
| Iterations | 1000 (fixed) |

Table 9: Hyperparameters used for tuning with Optuna for random forest

| Hyperparameter | Range/Options |
|---|---|
| n_estimators | 50 to 300 |
| max_depth | 2 to 32 (log scale) |
| min_samples_split | 2 to 16 |
| min_samples_leaf | 1 to 16 |
| max_features | {'sqrt', 'log2', None} |
| bootstrap | {True, False} |
| criterion | {'gini', 'entropy'} |
| class_weight | {None, 'balanced', 'balanced_subsample', 'custom'*} |
| max_samples | 0.5 to 1.0 (if bootstrap is True) |

\* For custom class weight, 1 to 11

*b. Best hyperparameters found*

Table 10: Best hyperparameters found using Optuna for CatBoost

| Hyperparameter | Value |
|---|---|
| Learning rate | 0.003 |
| Depth | 4 |
| Subsample | 0.398 |
| Colsample by level | 0.605 |
| Minimum data in leaf | 78 |
| Class weight (1) | 11.957 |
| Iterations | 1000 (fixed) |

Table 11: Best hyperparameters found using Optuna for random forest

| Hyperparameter | Value |
|---|---|
| n_estimators | 153 |
| max_depth | 8 |

| min_samples_split | 2 |
| min_samples_leaf | 12 |
| max_features | sqrt |
| bootstrap | False |
| criterion | entropy |
| class_weight | balanced |
| max_samples | 0.963 |

D. **Optional thresholds**

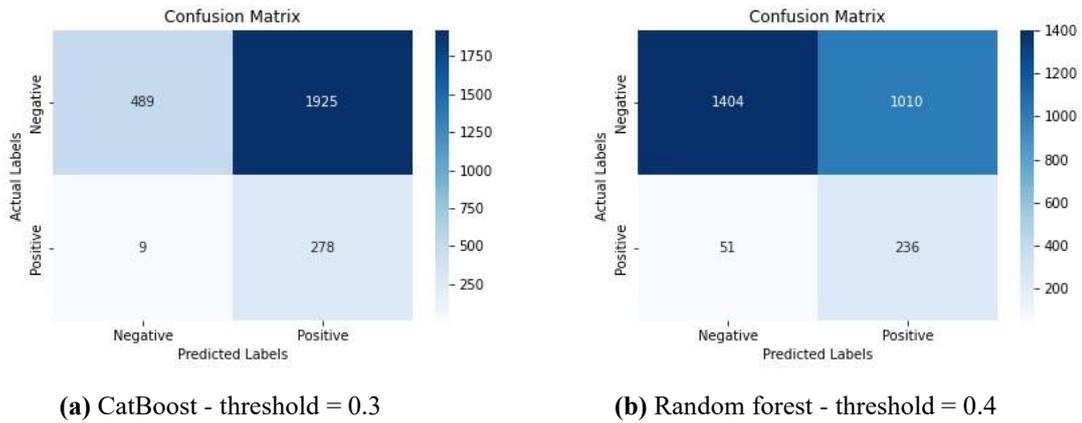

(a) CatBoost - threshold = 0.3    (b) Random forest - threshold = 0.4

**Figure 7:** Comparison of confusion matrices for CB and RF models with optional thresholds.

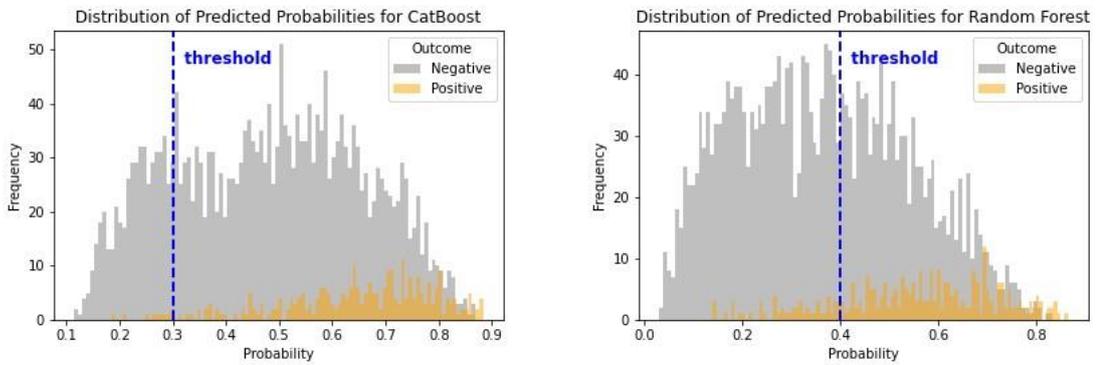

(a) CatBoost - threshold = 0.3 (b) Random forest - threshold = 0.4 **Figure 8:** Comparison of probability thresholds for CB and RF model.

E. **T-test results**

**Table 12:** Results of Mann-Whitney U test for blood culture outcome with pvalues (p < 0.05). Asterisk (*) indicates a not significant p-value.

| Variable | P-Value |
|---|---|
| Age | 3.63132e-32 |
| Sex | 2.68946e-09 |
| Alkaline_phosphatase | 6.4138e-37 |
| Basophils | 3.10109e-05 |
| Bilirubin | 2.47765e-89 |
| Creatinine | 1.98531e-31 |
| CRP | 1.13412e-23 |
| Eosinophils | 2.77477e-50 |
| Gamma_GT | 4.76017e-39 |
| Glucose | 7.87327e-26 |
| Hemoglobin | 1.31397e-08 |
| Hematocrit | 1.68843e-10 |
| Leukocytes | 1.36369e-25 |
| Lymfocytes | 5.33583e-120 |
| Monocytes | 4.30952e-11 |
| Neutrophils | 4.66592e-52 |
| Potassium | 0.00431216 |
| Sodium | 2.65448e-06 |
| Thrombocytes | 1.64344e-22 |
| Urea | 5.87107e-34 |
| Heartrate | 9.37253e-12 |
| Systolic_blood_pressure | 8.24811e-14 |
| Diastolic_blood_pressure | 1.0586e-31 |
| Temperature | 5.57477e-49 |
| Respiratory_rate | 0.303355* |
| Saturation | 0.0474665 |